\documentclass[IEEEtran]{journal}
\ifCLASSINFOpdf
  \usepackage[pdftex]{graphicx}
  \graphicspath{{../pdf/}{../jpeg/}}
\else
\fi
\usepackage[cmex10]{amsmath}
\usepackage{algorithm}
\usepackage[misc,geometry]{ifsym} 
\usepackage{amsmath}
\usepackage{amssymb}
\usepackage{mathtools}
\usepackage{verbatim}
\usepackage{graphicx}
\usepackage[dvipsnames]{xcolor}

\usepackage{multirow}
\usepackage{algpseudocode}

\usepackage{array}
\usepackage{mdwmath}
\usepackage{mdwtab}
\usepackage{eqparbox}
\usepackage[tight,footnotesize]{subfigure}
\usepackage[font=footnotesize]{subfig}

\DeclareUnicodeCharacter{2212}{-}
\DeclareUnicodeCharacter{2061}{}

\begin{document}
%
% paper title
% can use linebreaks \\ within to get better formatting as desired
\title{Neural Graph Matching for Modification Similarity Applied to Electronic Document Comparison}
%
%
% author names and IEEE memberships
% note positions of commas and nonbreaking spaces ( ~ ) LaTeX will not break
% a structure at a ~ so this keeps an author's name from being broken across
% two lines.
% use \thanks{} to gain access to the first footnote area
% a separate \thanks must be used for each paragraph as LaTeX2e's \thanks
% was not built to handle multiple paragraphs
%

\author{Po-Fang Hsu, Chiching Wei
\thanks{Po-Fang Hsu is with the Department of Advanced Document Technologies, Foxit Software Inc., 41841 Albrae Street. Fremont, CA, 94538 USA (e-mail: daniel\_hsu@foxitsoftware.com).}
\thanks{Chiching Wei (CTO) is with Foxit Software Inc., 41841 Albrae Street., Fremont, CA, 94538 USA (e-mail: jeremy\_wei@foxitsoftware.com).} 
}

% note the % follow\{ing the las, Chiching Wei\}t \IEEEmembership and also \thanks - 
% these prevent an unwanted space from occurring between the last author name
% and the end of the author line. i.e., if you had this:
% 
% \author{....lastname \thanks{...} \thanks{...} }
%                     ^------------^------------^----Do not want these spaces!
%
% a space would be appended to the last name and could cause every name on that
% line to be shifted left slightly. This is one of those "LaTeX things". For
% instance, "\textbf{A} \textbf{B}" will typeset as "A B" not "AB". To get
% "AB" then you have to do: "\textbf{A}\textbf{B}"
% \thanks is no different in this regard, so shield the last } of each \thanks
% that ends a line with a % and do not let a space in before the next \thanks.
% Spaces after \IEEEmembership other than the last one are OK (and needed) as
% you are supposed to have spaces between the names. For what it is worth,
% this is a minor point as most people would not even notice if the said evil
% space somehow managed to creep in.

% The paper headers
\markboth{Journal of \LaTeX\ Class Files,~Vol.~6, No.~1, January~2007}%
{Shell \MakeLowercase{\textit{et al.}}: Bare Demo of IEEEtran.cls for Journals}
% The only time the second header will appear is for the odd numbered pages
% after the title page when using the twoside option.
% 
% *** Note that you probably will NOT want to include the author's ***
% *** name in the headers of peer review papers.                   ***
% You can use \ifCLASSOPTIONpeerreview for conditional compilation here if
% you desire.

% If you want to put a publisher's ID mark on the page you can do it like
% this:
%\IEEEpubid{0000--0000/00\$00.00~\copyright~2007 IEEE}
% Remember, if you use this you must call \IEEEpubidadjcol in the second
% column for its text to clear the IEEEpubid mark.

% use for special paper notices
%\IEEEspecialpapernotice{(Invited Paper)}

\maketitle
\thispagestyle{empty}

\begin{abstract}
%\boldmath
This paper presents a novel neural graph matching approach applied to document comparison, which is a common task in legal and financial industries. In some cases, the most important differences are the addition or omission of words, sentences, clauses, or paragraphs. However, comparing document is a challenging task if the whole editing process is not recorded or traced. Under many temporal uncertainties, the potentiality of our approach is explored to approximate the accurate comparison to make sure which element blocks have a relation of edition with others. In the beginning, a document layout analysis that combines traditional and modern techniques to segment layout in blocks of various types is applied. Then this issue is transformed to a problem of layout graph matching with textual awareness. Graph matching is a long-studied problem with a broad range of applications. Different from previous works that focus on visual images or structural layout, we also bring textual features into our model to adapt this textually-rich domain. Specifically, based on the decodable electronic document, an encoder is introduced to deal with the visual presentation decoded from PDF. Additionally, since the modifications can cause the inconsistency of document layout analysis between modified documents and the blocks can be merged and split, Sinkhorn divergence is adopted in our graph neural approach, which tries to overcome both these issues with many-to-many block matching. This is demonstrated on two categories of layouts, legal agreement and scientific articles, collected from our real-case datasets.
\end{abstract}
% IEEEtran.cls defaults to using nonbold math in the Abstract.
% This preserves the distinction between vectors and scalars. However,
% if the journal you are submitting to favors bold math in the abstract,
% then you can use LaTeX's standard command \boldmath at the very start
% of the abstract to achieve this. Many IEEE journals frown on math
% in the abstract anyway.

% Note that keywords are not normally used for peerreview papers.

\IEEEpeerreviewmaketitle

\section{Introduction}
% The very first letter is a 2 line initial drop letter followed
% by the rest of the first word in caps.
% 
% form to use if the first word consists of a single letter:
% \IEEEPARstart{A}{demo} file is ....
% 
% form to use if you need the single drop letter followed by
% normal text (unknown if ever used by IEEE):
% \IEEEPARstart{A}{}demo file is ....
% 
% Some journals put the first two words in caps:
% \IEEEPARstart{T}{his demo} file is ....
% 
% Here we have the typical use of a "T" for an initial drop letter
% and "HIS" in caps to complete the first word.
\IEEEPARstart{D}{ocument} comparison, also known as redlining or blacklining, is a process of cross checking new versions of a document against previous copies in order to identify changes. These differences could include minor formatting modifications, such as font or spacing changes, or more significant grammatical changes. Document comparison is a common task in the legal and financial industries. In some cases, however, the most important differences may be the addition or omission of words, sentences, clauses, or paragraphs. 

Traditionally, document comparison is generally processed through three stages. The first stage is document layout analysis, which analyzes document structures using primitive heuristics, statistical methods, or machine learning. Block-matching is then applied to match all of the blocks according to their modification similarity. Finally, edit distance can be used to evaluate the string similarity for matching all of the text in each block pair.

\begin{figure}[!t]
\centering
\includegraphics[width=3.5in]{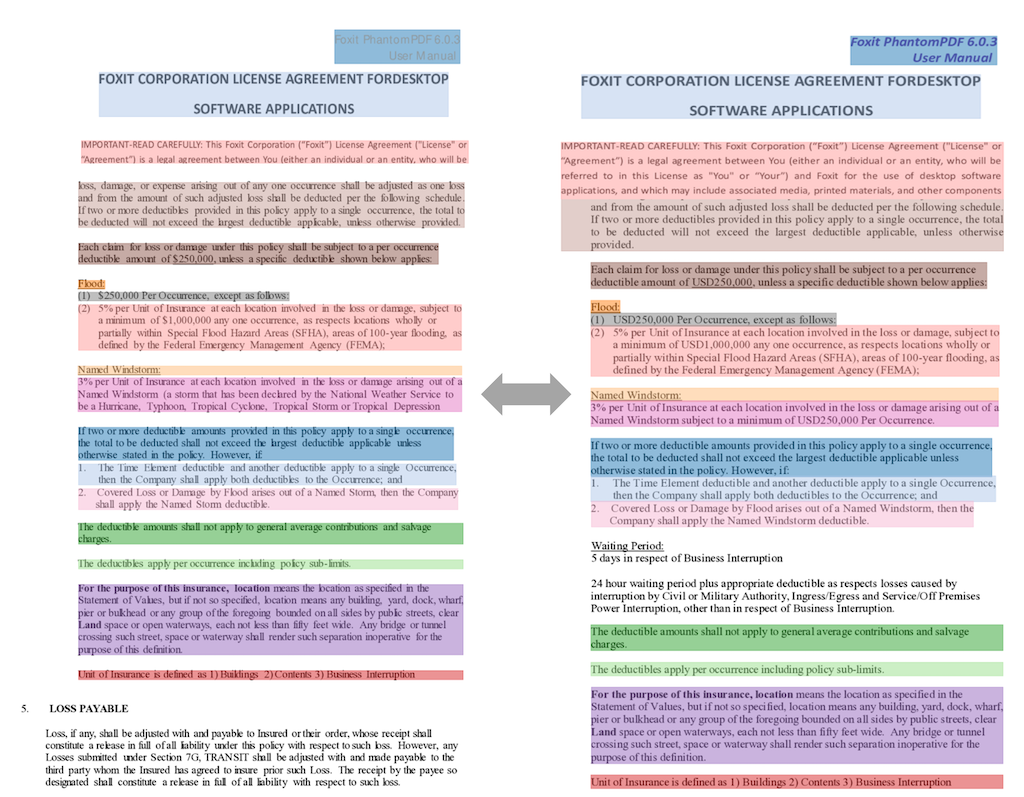}
\caption{Block Matching.}
\label{fig:block_matching}
\end{figure}

In the beginning, a document layout analysis process is applied that combines traditional and modern techniques to segment layout in blocks of various types. For example, a table should be classified into cells and another blocks can be paragraph, title, section, and so on. 
To perform a suitable comparing process, understanding the pairs of matched blocks should be compared is an indispensable step, such as in Fig.\ref{fig:block_matching}. However, this is not an easy task if the whole editing process is not recorded or traced. Under many temporal uncertainties, to make sure which blocks have a relationship of edition with each other, the best solution is not existed. For example, we cannot determine definitely whether a paragraph "123xxxx" is modified from paragraph "xx123oo", "123xx", or a completely new paragraph. Generally, the challenges can be classified into the following categories.

\paragraph{Semantic Inconsistency}
One type of modification is semantic inconsistency of the component. After modifying, a paragraph block can become a list item or another type of layout. The context of the text, nearby blocks, and style of blocks should be considered.

\begin{figure}[htbp]
\centerline{\includegraphics{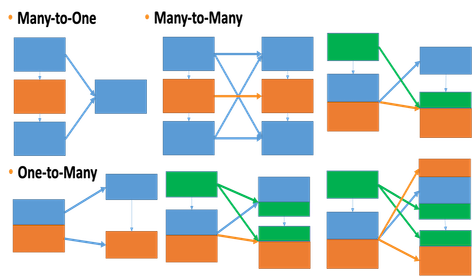}}
\caption{Matching Cases.}
\label{fig:matching_cases}
\end{figure}

\paragraph{Geometric Inconsistency}
The position of the component that is changed is called Geometric Inconsistency. This means that a component is shifted by reading order. Additionally, a block can be split into multiple blocks, or merged with other blocks as shown in  Fig.~\ref{fig:matching_cases}.

\paragraph{Text Ambiguity}
Because of the modification, it is difficult to distinguish which blocks should be paired in some situation. For example, a list item block “c) Ambiguous Text:” should be paired with list item block “c) Text Ambiguity:” or the other near list item block “d) Ambiguous Text:”.

An algorithm is designed to approximate the best solution. First, locate templates from two different versions with graph matching method. The graphs are then processed to find the most similar pairs in text level edited distance. 

Graph matching – establishes correspondences between two graphs represented in terms of both local node structure and pair-wise relationships, whether they are visual, geometric, or topological. It is important in areas such as combinatorial optimization, machine learning, image analysis, or computer vision, and has applications in structure-from-motion, object tracking, 2d and 3d shape matching, image classification, social network analysis, autonomous driving, and more.

 A novel two-stage neural approach, which overcomes both limitations. The first stage learns a representation for the layout in the page by combining both the text and visual information extracted from PDF code. This stage is able to largely generalize document format for acquiring all representation without expensive computing and storage features over visual renderings of the page. The second stage detects edited relatedness using a graph neural network. Unlike other traditional methods such as iterative closest point, which are limited to rigid displacements, graph matching naturally encodes structural information that can be used to model complex relationships and more diverse transformations.

\section{Related Works}\label{III}

\subsection{Document Comparison}

Document comparison\cite{b6, b7}, also known as redlining or blacklining, is a process of cross checking new versions of a document against previous copies in order to identify changes.

\subsection{Document Layout Presentation}

The recent work by Manandhar et al. \cite{b4} is the first to leverage GNNs to learn structural similarity of 2D graphical layouts, focusing on UI layouts with rectangular boundaries. The work employed a GCN-CNN architecture on a graph of UI layout images, also under an IoU-trained triplet network \cite{b19}. In addition, \cite{b9} learns the graph embeddings in a dependent manner. Through cross-graph information exchange, the embeddings are learned in the context of the anchor-positive (respectively, the anchor-negative) pair.

\subsection{Graph Matching}

\paragraph{Non-deep learning methods}
 The structural SVM based supervised learning method \cite{b15} incorporates earlier graph matching learning methods \cite{b0, b1, b2, b3}. Learning can also be fulfilled by unsupervised \cite{b2} and semi-supervised \cite{b16} mechanisms. In these earlier works, no neural network is adopted until the recent seminal work \cite{b5, b14}.

\paragraph{Deep-learning methods} Deep learning is recently applied for graph matching on images \cite{b14}, whereby convolutional neural network (CNN) is used to extract node features from images followed with spectral matching and CNN is learned using a regression-like node correspondence supervision. This work is improved by introducing GNN to encode structural \cite{b13} or geometric \cite{b18} information, with a combinatorial loss based on cross-entropy loss, and Sinkhorn network \cite{b17}. LayoutGMN \cite{b9} learns to predict structural similarity between
two layouts with an attention-based graph matching network.

\section{Our Approach}

\subsection{Decoding-Aware Visual Feature Presentation}

\begin{figure}[htp]
\centerline{\includegraphics{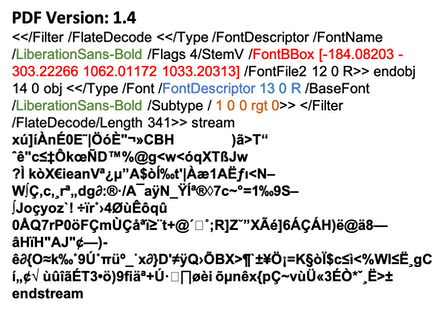}}
\caption{PDF Code.}
\label{fig:pdf_code}
\end{figure}

Previous general approaches have need of the image examples for extracting and aligning the visual information. Indeed, image is a significant important feature in document representations. However, for the time being, it is a huge surge in document electronically, almost formats of electronic document render with the same encoding. As PDF, displaying the exact same content and layout no matter which operating system, device or software application it is viewed on. For PDF 1.4, the visual features are embedded as shown in Fig.~\ref{fig:pdf_code}; font style is shown in (\textcolor{OliveGreen}{Green}), font size is shown in (\textcolor{RoyalBlue}{Blue}), color is shown in (\textcolor{Tan}{Tan}), position is shown in (\textcolor{Red}{Red}) and text encode is shown in (\textbf{Bold}).

\begin{figure*}[htbp]
\centering
\includegraphics[width=\textwidth]{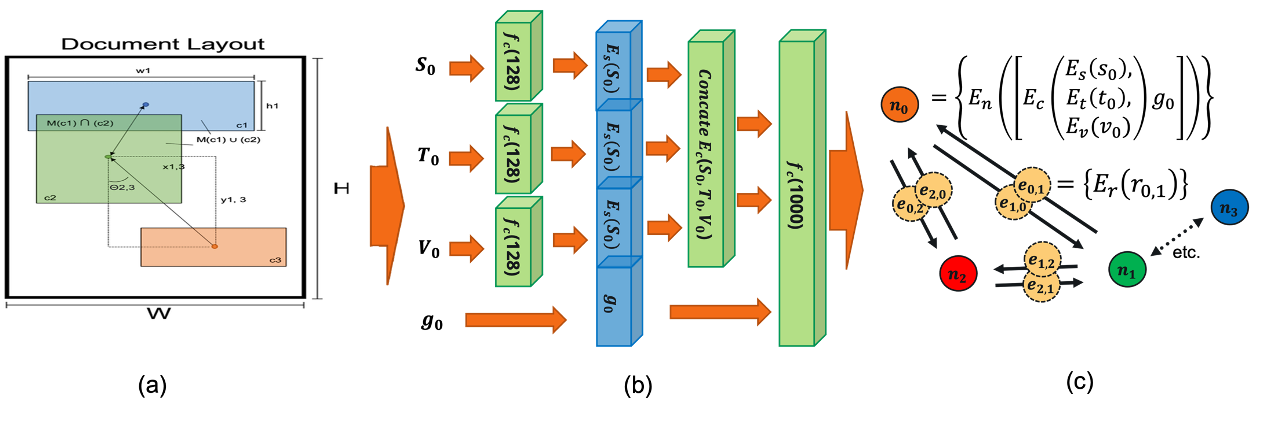}
\caption{Graph Encoder.}
\label{fig:graph_encoder}
\end{figure*}

\subsection{Graph Representation For Layout}
A variety of visual and content-based features can be incorporated as the initial node features, such as the text, font, size, and type of a layout element or the image features in a document page. For modified learning in decodable electronic document tasks such as ours, we can only focus on content-based features. However, our work is not similar to \cite{b8,b9}, the initial node features only contain semantic and geometric information of the layout elements. Since text and related features need to be considered, we simply apply Sentence-BERT\cite{b10} here to embed all of the text in the element. As shown in Fig.~\ref{fig:graph_encoder}(a), which describes a document layout with height $h$ and width $w$ as a spatial graph $G=(V,E)$ where $V=\{c_1, \cdots, c_i, \cdots, c_k\}$ is set of nodes representing its k components and $E=\{e_1, \cdots, e_i, \cdots, e_k\}$ is the set of edges that denotes a relationship between them. Each node carries three or four types of information:
	\begin{itemize}
	\item (Optional) Semantic property $s_i$: A one-hot vector denoting the component class.
	\item Text property $t_i$: All the embedded text inside the component.
	\item Geometric property $g_i$: Captures the spatial location of the component in layout are encoded.
	\item Visual property $v_i$: Extracts the visual features from PDF code.
    \end{itemize}

%
%\begin{figure}[ht]
%\centerline{\includegraphics{layout_representation_new.png}}
%\caption{Geometric property Representation.}
%\label{fig:layout_representation}
%\end{figure}

Let $(x_i,y_i)$ and $(w_i,h_i)$ be the geometrical centroid, width and  height of the component $c_1$, and $A_i=\sqrt{w_ih_i}$, then the geometric feature $g_i$ is $[\frac{x_i}{w},\frac{y_i}{h},\frac{w_i}{w},\frac{h_i}{h},\frac{A_i}{wh}]$.

Define the edges features $r_{ij}$ associated with edge $e_{ij}$ using the pairwise geometric features between components $c_i$ and $c_j$ given below.
\begin{equation}
r_{ij}=[\phi_{ij},\frac{w_j}{w_i} ,\frac{h_j}{h_i} ,\frac{1}{D} \sqrt{\Delta x^2 + \Delta y^2}, \frac{\Delta x }{A_i}, \frac{\Delta y}{A_i} , \theta_{ij}] \label{eq:edges_features}
\end{equation}
Where $\Delta x=x_j+x_i$ and $\Delta y=y_j+y_i$ are the shifts between the components and constant 
$D=\sqrt{w^2+h^2}$ normalises against the diagonal. In addition, the feature $r_{ij}$  incorporates various geometric relations such as relative distance, aspect ratios and orientation
 $\theta=\arctan 2(⁡\frac{\Delta y}{\Delta x}) \in [-\pi,\pi]$. 
 We explicitly include a containment feature $\psi_{ij}$ taking into account the Intersection over Union (IoU) between components capturing the nesting of the layout components:

\begin{equation}
\psi_{ij}=\frac{(M(c_i) \cap M(c_j))}{(M(c_i) \cup M(c_j)}\label{eq:iou}
\end{equation}

 The visual feature $v_i$ is a 3xN matrix of which N is the same as the length of the text of component, and the 3 types of features (Font-style, Size, and Color) are obtained from PDF code:
	\begin{itemize}
	\item Font-style encodes typeface and font. For example, LiberationSans-Bold is encoded as below. 
          \begin{center}
                \begin{tabular}{ |c|c|c| } 
                 \hline
                  LiberationSans & $\cdots\cdots$ & Bold \\
                 \hline
                 1 & 0 & 1 \\
                 \hline
                \end{tabular}
                \end{center}
	\item Size is the size of text.
	\item Color is obtained by transforming color space to YCbCr with 4:2:0 chroma subsampling.
    \end{itemize}

%\begin{figure}[]
%\centerline{\includegraphics{layout_encoder.png}}
%\caption{Graph Representation Layout.}
%\label{fig:layout_encoder}
%\end{figure}

\subsection{Layout Graph Encoder}
As show in Fig.~\ref{fig:graph_encoder}(c), the node features $n_i$ in the graph hold both the semantic class label $s_i$ as well as the geometric property $g_i$, text property $v_i$ and visual presentation $v_i$ of the layout component $c_i$. 
$n_0 = {E_n([E_c(E_s(s_0),E_t(t_0),E_v(v_0)) g_0])}$
where $E$ is the embedding layer that learns the several kinds of features and then projects the features into node feature $n_i$. Similarly, the edge features $r_{ij}$ are projected by $E_r(r_{ij})$. Next, the node features and the edge (relation) features are operated by graph convolutional networks
$g_n (\cdot)$ and $g_r (\cdot)$. The node and relational feature outputs of the GCN network are computed by
$x_{n_i}=g_n (n_i)$ , $x_{r_{ij}}=g_r([n_i E_r (r_{ij})n_j])$.
The relation graph network $g_r$ operates on tuples $\langle n_i, E_r (r_{ij} )$, $n_j \rangle$ passing the information through the graph to learn the overall layout. Then obtain two set of features  $X_n=\{x_{n_1},x_{n_2},\cdots,x_{n_n}\}, X_r=\{x_{r_1},x_{r_2},\cdots,x_{r_n}\}$. Where $k$ and $k^{'}$ are the number of components (node features) and the total number of the relationship features which vary for different layout layouts. Next, the sets of features are passed through self-attention modules which learn to pool the node features and relational features given by

\begin{equation}
f_{n}^{att}=\sum_{i=1}^k a_{n_i} x_{n_i}
,
f_{r}^{att}=\sum_{i=1}^{k^{'}} a_{r_i} x_{r_i}\label{eq:node_features}
\end{equation}

\begin{equation}
\frac{\exp (w_{n}^{\intercal} x_{n_i})}{\sum_{l=1}^{n} \exp (w_{n}^{T}x_{n_l})}
,
\frac{\exp (w_{r}^{\intercal} x_{r_i})}{\sum_{l=1}^{k^{'}} \exp (w_{r}^{T}x_{r_l})}\label{eq:relational_features}
\end{equation}

where, $a_{n_i}$ and $a_{r_i}$ are attention weights learned with $w_{n}^{\intercal}$ and $w_{r}^{\intercal}$ parameters.

Finally, obtain a d−dimensional latent embedding that encodes the layout:
\begin{equation}
f_e=E_e ([f_{n}^{att},f_{r}^{att}])
\label{eq:layout_encodes}
\end{equation}

\subsection{Layout Graph Matching Neural Network}
We modify previous works \cite{b11, b12, b13} to implement our approach.
The learning process of our matching neural network is shown in Alg.~\ref{alg:cross-graph}.

\begin{algorithm}[htbp]
  \caption{Iterative cross-graph node embedding}\label{alg:cross-graph}
  \label{alg:ns}
  \begin{algorithmic}[1]
    \Require
      Block features $\{ b_{1i}^{(0)},b_{2j}^{(0)} \}_{i \in \nu_1,j \in \nu_2}$;
      number of iterations K
    \Ensure
      embedding features $\{ b_{1i}^{(3)},b_{2j}^{(3)} \}_{i \in V_1,j \in V_2}$
    \State // first intra-graph aggregation
    \State $\{ b_{si}^{(1)} \} \leftarrow GConv_1(A_s, \{h_{si}^{(0)} \});$
    \State // Initialize $\hat{S}^{(0)}$ as zero matrix
    \State $\hat{S}^{(0)} \leftarrow 0^{N \times N}$;
    \For{$k \leftarrow \{1...K\}$}
    \State // cross-graph aggregation
    \State $\{h_{1i}^{(2)} \} \leftarrow GConv(\hat{S}^{(k-1)}\{h_{1i}^{(1)}\}, \{h_{2i}^{(1)}\});$
    \State $\{h_{1i}^{(2)} \} \leftarrow GConv(\hat{S}^{(k-1)\top}, \{h_{2i}^{(1)}\}, \{h_{1i}^{(1)}\});$
    \State // second intra-graph aggregation
    \State $\{h_{si}(3) \} \leftarrow GConv(A_s, \{h_{si}^{(2)}\});$
    \State // correspondence prediction 
    \State build $\hat{M}$ from $\{h_{1i}^{(3)}\}, \{h_{2i}^{(3)}\};$
    \State $\hat{S}^{(k)} \leftarrow Sinkhorn(M);$
    \EndFor
  \end{algorithmic}
\end{algorithm}
    
\section{Datasets}

To the best of our knowledge, this work is the first to explore the topic of document comparison in  visually-rich documents. Two kinds of layout datasets are collected and constructed: legal agreement and scientific articles. After data filtering, the size of the two datasets is 30 and 25 respectively. The component-level region bounding box is provided according to manual validation for the prediction of our Cascade Mask R-CNN\cite{b20} based document layout analysis detector. The corresponding transcript is annotated with the help of in-house workers, using a Web based annotation tool, Label Studio\cite{b21}. The annotation process was organized into two stages. In Stage 1, workers were shown a pair of document images with the annotation of layout segmentation and they revised the wrong annotation predicted from DLA predictor. Stage 2 aims to construct the matching relation between the components. Workers simply used the Relations tag provided by Label Studio to create label relations between regions.

\section{Conclusions}

A novel graph neural network is presented to solve both context similarity of components and structural matching between layout components. The main limitation of the current proposed framework is the requirement for strong a priori knowledge, which means that the framework is not so robust in revising the wrong input from document layout analysis. Another limitation of the current network is that it does not learn hierarchical graph representations which is desirable when dealing with cross-page comparison.

\ifCLASSOPTIONcaptionsoff
  \newpage
\fi


\begin{thebibliography}{1}
\bibitem{b0} L. Torresani, V. Kolmogorov, and C. Rother, “Feature correspondence via graph matching: Models and global optimization,” in
Eur. Conf. Comput. Vis. Springer, 2008, pp. 596–609.
\bibitem{b1} T. Caetano, J. McAuley, L. Cheng, Q. Le, and A. J. Smola, “Learning graph matching,” Trans. Pattern Anal. Mach. Intell., vol. 31, no. 6, pp. 1048–1058, 2009.
\bibitem{b2} M. Leordeanu, R. Sukthankar, and M. Hebert, “Unsupervised learning for graph matching,” Int. J. Comput. Vis., vol. 96, no. 1, pp. 28–45, 2012.
\bibitem{b3} Andrei Zanfir, “Deep Learning of Graph Matching,” IEEE/CVF Conference on Computer Vision and Pattern Recognition, 2018.
\bibitem{b4} Dipu Manandhar, Dan Ruta, and John Collomosse, “Learning Structural Similarity of User Interface Layouts using Graph Networks,” ECCV, 2020.
\bibitem{b5} Minsu Cho, “Progressive Graph Matching: Making a Move of Graphs via Probabilistic  Voting, CVPR.2012.
\bibitem{b6} Andrew Kao, “How to compare two PDF documents side by side,” https://www.foxitsoftware.com/blog/how-to-compare-two-pdf-documents-side-by-side/
\bibitem{b7} ADOBE ACROBAT, “Easily compare PDF files.,” https://acrobat.adobe.com/us/en/acrobat/how-to/compare-two-pdf-files.html
\bibitem{b8} Dipu Manandhar et al, “Learning Structural Similarity of User Interface Layouts using Graph Networks,” CVPR 2021.
\bibitem{b9} Akshay Gadi Patil et al, “LayoutGMN: Neural Graph Matching for Structural Layout,” CVPR 2021.
\bibitem{b10} Nils Reimers and Iryna Gurevych, “Sentence-BERT: Sentence Embeddings using Siamese BERT-Networks,” ACL 2019.
\bibitem{b11} R Wang, J Yan, X Yang, “Learning combinatorial embedding networks for deep graph matching,” The IEEE International Conference on Computer Vision (ICCV), October 2019.
\bibitem{b12} Wang, J Yan, X Yang, “Combinatorial Learning of Robust Deep Graph Matching: an Embedding based Approach,” IEEE Transactions on Pattern Analysis and Machine Intelligence, 2020.
\bibitem{b13} R. Wang, J. Yan and X. Yang, “Neural Graph Matching Network: Learning Lawler’s Quadratic Assignment Problem with Extension to Hypergraph and Multiple-graph Matching,” TPAMI 2021
\bibitem{b14} A. Zanfir and C. Sminchisescu, “Deep learning of graph matching,” Comput. Vis. Pattern Recog., 2018, pp. 2684–2693.
\bibitem{b15} M. Cho, K. Alahari, and J. Ponce, “Learning graphs to match,” Int. Conf. Comput. Vis., 2013, pp. 25–32.
\bibitem{b16} M. Leordeanu, A. Zanfir, and C. Sminchisescu, “Semi-supervised learning and optimization for hypergraph matching,” Int. Conf. Comput. Vis. IEEE, 2011, pp. 2274–2281.
\bibitem{b17} R. Adams and R. Zemel, “Ranking via sinkhorn propagation,” arXiv:1106.1925, 2011.
\bibitem{b18} Z. Zhang and W. S. Lee, “Deep graphical feature learning for the feature matching problem,” Int. Conf. Comput. Vi., 2019, pp. 5087–5096.
\bibitem{b19} Elad Hoffer and Nir Ailon, “Deep metric learning using triplet network,” International Workshop on Similarity-Based Pattern Recognition, pages 84–92. Springer, 2015.
\bibitem{b20} Junlong Li et al., “DiT: Self-supervised Pre-training for Document Image Transformer,” arXiv 2022.
\bibitem{b21} Maxim Tkachenko et al., “{Label Studio}: Data labeling software,” 2020-2021.
\end{thebibliography}
\end{document}